\documentclass[letterpaper, 10 pt, conference]{ieeeconf}  
\IEEEoverridecommandlockouts                              
\usepackage[nomessages]{fp}
\usepackage{graphicx}
\usepackage{amsmath}
\usepackage{amsfonts}
\usepackage[caption=false,font=footnotesize]{subfig}
\usepackage{caption}
\usepackage{tabularx}
\usepackage{multirow}
\usepackage{cite}
\usepackage{booktabs}
\usepackage{textcomp}
\usepackage[symbol]{footmisc}
\usepackage{float}
\usepackage{makecell}
\usepackage{ntheorem}
\theoremseparator{:}
\newtheorem{hyp}{Hypothesis}
\usepackage[binary-units=true]{siunitx}

\title{\LARGE \bf
Comparing Alternate Modes of Teleoperation for Constrained Tasks
}

\author{Christopher E. Mower$^{1}$, Wolfgang Merkt$^{1}$, Aled Davies$^{2}$, and Sethu Vijayakumar$^{1}$%
  \thanks{$^{1}$School of Informatics, University of Edinburgh, UK.
    Email: {\tt\small chris.mower@ed.ac.uk}}%
  \thanks{$^{2}$Infrastructure division, Costain PLC, UK.}%
  \thanks{This research is supported by the Engineering and Physical Sciences Research Council (EPSRC, grant reference EP/L016834/1) and EU H2020 project Memory of Motion (MEMMO, project ID: 780684). Christopher E. Mower is partially supported by The Costain Group PLC.}%
}

\newcommand{\hypref}[1]{{\bfseries H\ref{#1}}}

\begin{document}
\bstctlcite{IEEEexample:BSTcontrol}

\maketitle{}
\thispagestyle{empty}
\pagestyle{empty}

\begin{abstract}
Teleoperation of heavy machinery in industry often requires operators to be in close proximity to the plant and issue commands on a per-actuator level using joystick input devices. However, this is non-intuitive and makes achieving desired job properties a challenging task requiring operators to complete extensive and costly training.
Despite this, operator fatigue is common with implications for personal safety, project timeliness, cost, and quality. While full automation is not yet achievable due to unpredictability and the dynamic nature of the environment and task, shared control paradigms allow operators to issue high-level commands in an intuitive, task-informed control space while having the robot optimize for achieving desired job properties.

In this paper, we compare a number of modes of teleoperation, exploring both the number of dimensions of the control input as well as the most intuitive control spaces. Our experimental evaluations of the performance metrics were based on quantifying the difficulty of tasks based on the well known Fitts' law as well as a measure of how well constraints affecting the task performance were met. Our experiments show that higher performance is achieved when humans submit commands in low-dimensional task spaces as opposed to joint space manipulations.
\end{abstract}

\section{Introduction}
Teleoperation of industrial manipulators is generally repetitive and requires high levels of concentration and manual dexterity. Excessive cognitive loads invariably lead to fatigue that can become dangerous. This danger is prevalent in the construction sector seen by having one of the highest levels of incidents involving fatalities and serious injuries per annum in the United Kingdom~\cite{hse-stats-17}.


\begin{figure}[t]
  \centering
  \includegraphics[width=\columnwidth,trim={0cm 3cm 0cm 3cm},clip]{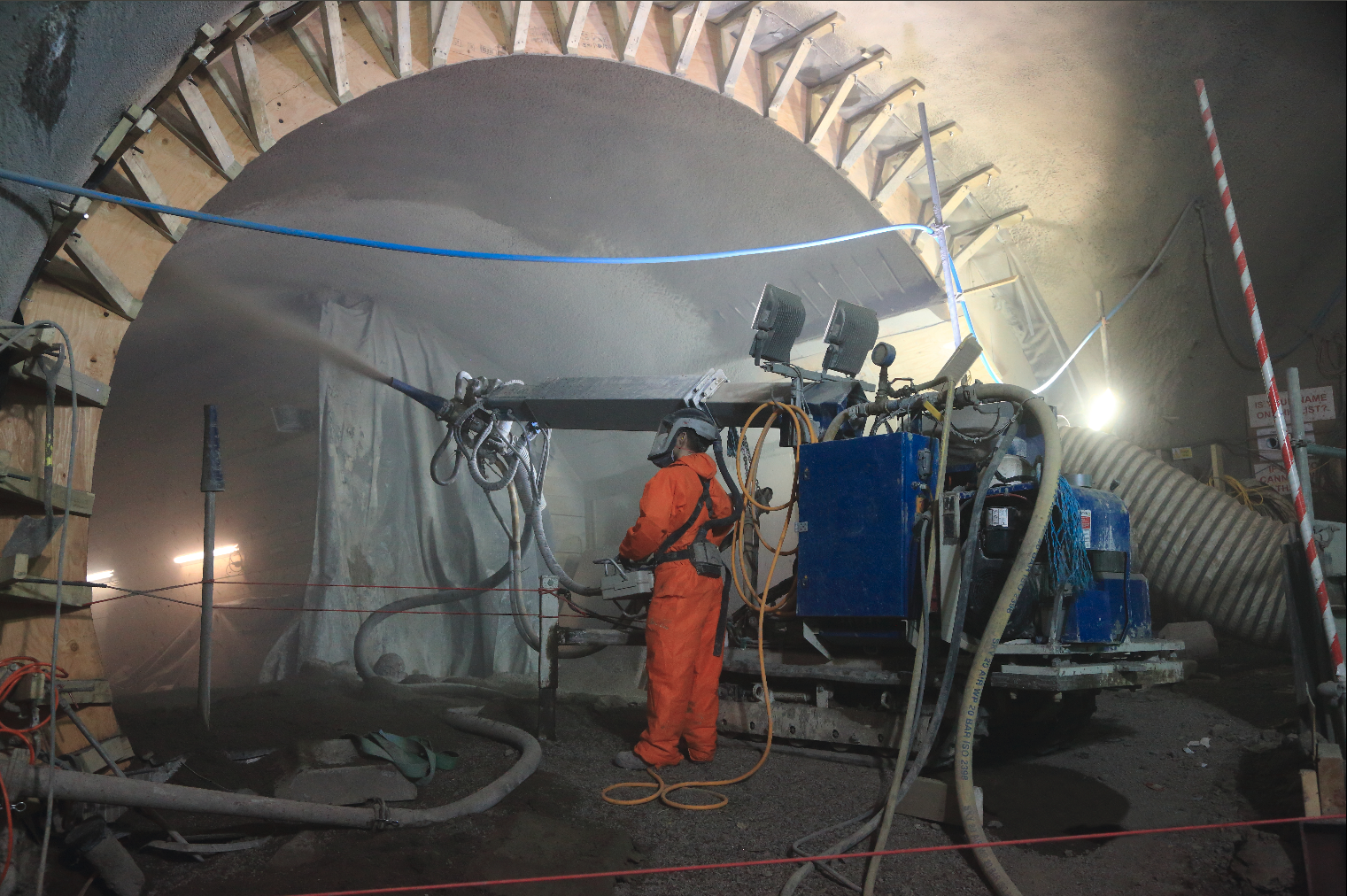}
  \caption{Shotcrete application in a freshly excavated tunnel using a 5-DoF concrete spraying unit. Image provided by Costain Laing O'Rourke Joint Venture.}
  \label{fig:oruga}
  \vspace{-1.5em}
\end{figure}


An example of such a task is concrete spraying as shown in Fig.~\ref{fig:oruga}. Here, a skilled human operator manipulates the device via some interface to spray a lining of wet concrete onto a excavated tunnel surface. The tunnel surface is often unstructured, due to excavation, and the operators visibility is restricted by high amounts of dust. Despite these restrictions, operators are required to manipulate these devices so that they (1) ensure job-site safety, (2) achieve high task performance, and (3) minimize costs~\cite{Ballou03ShotcreteSM}. Simultaneously accounting for safety and task performance in teleoperation tasks is required in a multitude of industries: robotic surgery~\cite{Sung01}, nuclear waste disposal~\cite{Farraj16}, space robotics~\cite{Lii10}, assembly~\cite{Sagardia16}, and subsea~\cite{Murphy11}.

Devices, such as the concrete sprayer systems, are generally controlled on a per-actuator level and consequently these control architectures do not lend themselves to easy operation since they force the operator to submit commands directly in the joint space. Since humans typically model tasks in the three dimensional Cartesian space they must learn naturally an inverse kinematic mapping imposing high costs in terms of monetary costs and time for specialized training. In the literature, methods such as the inverse-kinematics method \cite{Khatib87} and optimization techniques~\cite{Escande14, Zucker13} have been developed that can allow control commands to be submitted in alternative control spaces. A number of works leveraging these advancements have developed assistive techniques for teleoperation, for example, virtual fixtures~\cite{Rosenberg93VirtualVFP}, shared control frameworks that merge human input and autonomy~\cite{Farraj16}, and human supervisory capabilities implementing sliding autonomy~\cite{Merkt17}. However, for assistive techniques in general the issue of sub-task allocation between human and autonomy remains one of the main challenges~\cite{inagaki2003adaptive}. 

In this work, we investigate the allocation of sub-tasks in the context of control spaces for target acquisition tasks. That is, we ask which control spaces should a human operator submit commands in order to achieve high task performance? Our main source of influence comes from the concrete spraying example~\cite{Girmscheid01} however our investigation can be generalized to other tasks such as paint spraying~\cite{chen2002automated}, semi-autonomous grasping~\cite{Vogel16}, wiping~\cite{Armesto18}, and robotic surgery~\cite{ryden2012forbidden}. We have designed an experiment incorporating the well-known Fitts' law, that quantifies task difficulty, in order to compare and contrast different control modes. This paper provides knowledge of intuitive and effective control spaces for teleoperation and shared control that will allow for more grounded formulations. Our main contributions are listed as follows: 
\begin{enumerate}
    \item Analysis into the sub-task allocation for target acquisition tasks through an empirically driven investigation. 
    \item An extensive data set of results that contains objective and subjective metrics. 
    \item A comparison between two subsets of our participants based on the participants personal habits that we identified to effect performance. 
    \item Two generalized performance metrics for Fitts' law relating to teleoperated target acquisition tasks. 
\end{enumerate}

The resulting data support several conclusions about control, user preferences, and how habitual traits may effect task performance. Our analysis indicates that there is a relation between control space dimensionality and performance. It has been seen in our experiments that lower dimensional task spaces in general elicit highest performance and these are generally preferred by users. Users who were identified to play video games on a regular basis were generally able to complete tasks quicker than those who do not. However, those who do not play video games were seen to maintain other performance metrics to a higher standard. An accompanying video is available at \texttt{https://youtu.be/OLev3yawHqE}.

\section{Related Work}
Prior work has explored techniques that guide robots motions via synthesized constraints, i.e., potential fields and virtual fixtures. Potential fields, originally developed by Khatib~\cite{Khatib85}, guide a user towards or away from a goal or obstacle~\cite{Crandall02}. Virtual fixtures create virtual barriers or forbidden regions in the task space that provide assistance in protecting sensitive areas. Rosenberg~\cite{Rosenberg93VirtualVFP} provided early work in the development of virtual fixtures. A number of guidance and/or preventative virtual fixture designs were compared using a Fitts' law paradigm for a remote teleoperation peg placement task. Later work~\cite{Nia14} has developed on-the-fly forbidden region generation techniques using real-time sensor data. 

The work of Dragan and Srinivasa~\cite{Dragan12, Dragan13a} formalizes assistive teleoperation under a framework of policy blending. The system, grounded in inverse reinforcement learning, attempts to predict the intentions of the human operator in order to arbitrate the operator and robot control policies. Other learning-based methods have also been developed by Abi-Farraj et al.~\cite{AbiFarraj17} and attempt to refine unskilled operator input based on skilled operator input learned by exploiting learning from demonstration techniques. 

From the concrete spraying literature, a task space control framework was developed by Honegger and Codourey~\cite{Honegger98} based on the inverse-kinematics method. The system allows the operator to control the end-effector of the concrete spraying unit using a 6-DoF space mouse. Later work by Girmscheid and Moser~\cite{Girmscheid01} developed a sliding autonomy approach for a high-DoF concrete spraying unit. They define three levels of autonomy: (1) A manual mode, consisting of no autonomy and implementing common joint-level control, (2) A semi-automated mode that, using a pre-collected laser scan of a tunnel cavity and computed geometry, allows the operator to command only the position on the wall to spray whilst the system accounts for motion constraints, and (3) An automated mode that plans and executes an entire trajectory. The latter mode, they note, is only in development for tunnel-boring machine projects where the excavated tunnel surface is regular and very smooth. Modes of teleoperation compared in this work are inspired by these control techniques developed for concrete spraying applications.

The above works represent a number of developments in shared autonomy and collaborative manipulation. There exists in the literature a number of studies that compare various methods for different levels of task complexity. A number of strategies were compared by You and Hauser~\cite{You11} for a reach-to-target task that must deal with collision avoidance, dynamic constraints, and erroneous input. A human was given the task of controlling a robot arm to reach to a given target. The robot was positioned in a simulated environment of various different complexities. The strategies compared in their study each have varying levels of autonomy: direct joint control, inverse kinematics, inverse kinematics with predictive safety filter, reactive potential field, and sampling-based motion planner. Kim et al.~\cite{Kim12} compared manual and autonomous control modes for a pick-and-place task. A study by Leeper et al.~\cite{Leeper12} compares a number of strategies for assisted and non-assisted remote robot grasping. These studies compare various methods for varying levels of autonomy and/or interface designs. To the best of our knowledge, there has not been work in educing control spaces that are intuitive and elicit high task performance for unskilled operators. In this work, we use the well-known Fitts' law to quantify task complexity in order to systematically compare the performance of several modes of teleoperation. 

\section{Methodology}
\subsection{General formulation}
We assume a teleoperation setting akin to Fig. \ref{fig:oruga}, i.e., a human operates a manipulation system via some interface. Let us model the process of teleoperation by the functional
\begin{equation}\label{eq:teleoperation-model}
    x_{t+1} = f(x_t, h_t)
\end{equation}
where $x_t\in\mathcal{C}^n$ represents joint configuration for an $n$-DoF manipulator, $h_t\in\mathcal{H}^m$ are inputs from the interface, and $t$ denotes time. Note, we assume $\mathcal{C}^n$ and $\mathcal{H}^m$ to be bounded, with control inputs scaled to the range $[-1,1]$. Assuming a known initial state $x_0$, all future state can thus be computed by \eqref{eq:teleoperation-model}. It is also assumed that the function $f$ can be computed fast enough for real-time actuation. 

The chosen value for $m$ coupled with the space in which commands $h_t$ are submitted categorizes what we define as the \textit{mode of teleoperation}. In Section~\ref{sec:modes-of-teleoperation}, we describe four modes developed for our experiments to test our hypotheses detailed in Section~\ref{sec:hypothseses}. To simplify our notation, the remainder of the paper will neglect sub-script $t$.

\subsection{Hypotheses}\label{sec:hypothseses}
Below we state our hypotheses; these represent alternative hypotheses. Hypotheses \hypref{hyp:first} and \hypref{hyp:second} relate to the objective results (Section~\ref{sec:results-low-dim-task-high-perf}) whilst \hypref{hyp:third} and \hypref{hyp:fourth} relate to the subjective results (Section~\ref{sec:subjective-results}). As mentioned above, humans typically model tasks in three-dimensional Cartesian spaces as opposed to the manipulator joint configuration space and thus intuitively we expect that modes of teleoperation that allow control commands to be submitted in these task spaces will achieve higher performance. Therefore, we base our alternative hypotheses on this intuition. 

\begin{hyp}[H\ref{hyp:first}] \label{hyp:first}
Modes of teleoperation with lower dimensions will see higher performance.
\end{hyp}

\begin{hyp}[H\ref{hyp:second}] \label{hyp:second}
Modes of teleoperation that submit commands in the manipulator task space will see higher performance.
\end{hyp}

\begin{hyp}[H\ref{hyp:third}] \label{hyp:third}
Participants will rate higher those modes of teleoperation with fewer dimensions. 
\end{hyp}

\begin{hyp}[H\ref{hyp:fourth}] \label{hyp:fourth}
Participants will rate higher those modes of teleoperation that submit commands in the manipulator task space. 
\end{hyp}

\vspace{-1.75em}

\subsection{Modes of teleoperation}\label{sec:modes-of-teleoperation}

\subsubsection{Full joint mode (FJ)}
Industrial manipulators are operated on a per-actuator level. That is, there exists a one-to-one mapping between each joystick axis and manipulator actuator. Thus, the mode of teleoperation can be modeled using 
\begin{equation}\label{eq:mode-FJ}
    f(x, h) := x + \Delta t A h
\end{equation}
where $A$ is a  diagonal matrix such that the $n$ entries are maximum joint velocities, and $\Delta t$ is the control loop time step. Joint limits are handled by checking \eqref{eq:mode-FJ} prior to sending actuation signals. 

\subsubsection{Reduced joint mode (RJ)}
Here, we relieve a number of joints from the human and assign these as a task constraint. A split in the joints of the manipulator must be made; we split the manipulator in the middle relieving joints near the end-effector from the human control. Let $x = [x^{(1)}, x^{(2)}]^T$ where $x^{(1)}\in\mathcal{C}^m$ denote the joints under human control, and $x^{(2)}\in\mathcal{C}^{n-m}$ denote the autonomous joints. Inspired by the spraying a wall task, joints $x^{(2)}$ are assigned the task of end-effector alignment with the surface normal. We model the mode using 
\begin{equation}
    f(x, h) := [f^{(1)}(x, h), f^{(2)}(x)]^T
\end{equation}
where $f^{(1)}$ represents the next joint state for joints $1{:}m$ and similarly for $f^{(2)}$. The human controlled joints are computed in the same way as in the full joint control mode and thus, $f^{(1)}$ takes the same form as \eqref{eq:mode-FJ}. The functional form for $f^{(2)}$ is found using an unconstrained optimization expressed by
\begin{equation}
    f^{(2)}(x^{(2)}) := \underset{x^{(2)}\in\mathcal{C}^{n-m}}{\arg\min} ~ \big\|\phi_{so}(x^{(2)}) - y^* \big\|^2
\end{equation}
where $\phi_{so}(\cdot)$ is the mapping from the configuration space to the angular task space, $\|{\cdot}\|$ denotes the Euclidean norm, and $y^*$ represents the two-dimensional (pitch and roll angles) task space goal. 

\subsubsection{Full task mode (FT)}
An alternative class of modes is defined when the human instead submits commands to the manipulator task space rather than the joint space. For spraying, the manipulator task space comprises three translational and two rotational dimensions of the end-effector. This mode is thus expressed by
\begin{equation}
    f(x, h) := \underset{x\in\mathcal{C}^n}{\arg\min} ~ \big\| \phi(x) - h\big\|^2
\end{equation}
where $\phi(\cdot)$ is the mapping from the configuration space to the translational and angular task space. 

\subsubsection{Reduced task mode (RT)}
Humans typically model manipulation tasks in the task space. Spraying for an intuitive point of view is a two-dimensional task, i.e., the position on the wall to spray. Assuming some model of the environment geometry $e$, a two-dimensional point on the surface defined by the human $h$ can be used to generate task space goal. We model this mode by
\begin{equation}
     f(x, h)  := \underset{x\in\mathcal{C}^n}{\arg\min} ~ \|\phi(x) - y^*(h; e)\|^2
\end{equation}
where $y^*=y^*(h; e)$ is some five-dimensional task space goal computed using the two-dimensional input from the human. The computed value of $y^*$ encodes ideals such as perpendicularity to the spraying plane and a given standoff distance from the surface. For example, as in Fig. \ref{fig:metrics}, this amounts to minimizing $\theta_t$ and maintaining a constant value, $\delta^*$ say, for $\delta_t$.
\vspace{-1em}
\section{Experiments}
\subsection{Participant selection}\label{sec:participant-selection}
\begin{figure}
    \centering
    \includegraphics[width=0.35\textwidth, trim={4cm, 0cm, 0cm, 0cm}, clip]{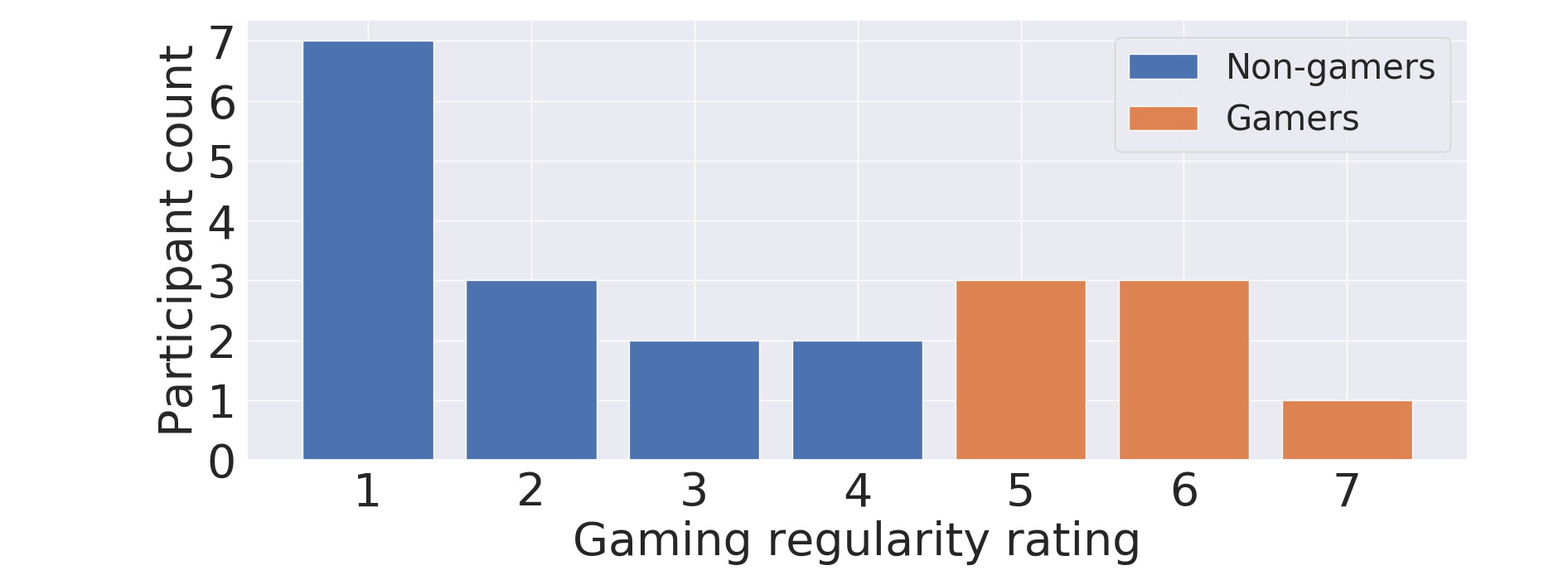}
    \caption{Participant selection gaming regularity.}
    \label{fig:participant-selection}
    \vspace{-2em}
\end{figure}

We have obtained results in this study by evaluating the performance of 21 participants (16 male, 5 female). The age distribution of the participants were 7 (21-25), 11 (26-30), 2 (31-35), and 1 (36+).  During preliminary investigations a difference in performance was noticed for participants who regularly played video games. During the final experiments, we asked participants to provide a rating on how often they play video games\footnote[2]{Note, those who participated in the preliminary investigation were barred from the final experiments.}. They were asked ``How often do you play video games (e.g., Xbox, PS4, PC)? [1 Never], [2 Bi-monthly], [3 Monthly], [4 Bi-weekly], [5 Weekly], [6 Regularly (but not daily)], [7 Daily]''. The responses are shown in Fig. \ref{fig:participant-selection}. A participant was considered a \textit{gamer} if they gave a rating greater than or equal to 4, and a \textit{non-gamer} otherwise.

\subsection{Experimental design}
Fitts' law is widely considered to be a robust measure of performance for target acquisition (pointing) tasks that makes the analogy between movement time and transmission of information~\cite{fitts1954information}. A common usage of Fitts' law from the human-computer interaction literature is to compare the usability of computer input devices (e.g. a mouse, trackball, and a stylus with a tablet)~\cite{MacKenzie91}. The law makes two underlying assumptions: (1) task difficulty is linearly correlated with performance, and (2) a complete move is performed through a number of iterations of feedback-guided corrective sub-movements, i.e. the \textit{deterministic iterative-corrections model}~\cite{Crossman83}. We use Fitts' law as a method to specify task difficulty. 

\begin{figure}[t]
  \centering
  \subfloat[An operator controlling the KUKA LWR arm.]{
    \includegraphics[width=0.25\textwidth,trim={8cm 12cm 12cm 10cm},clip]{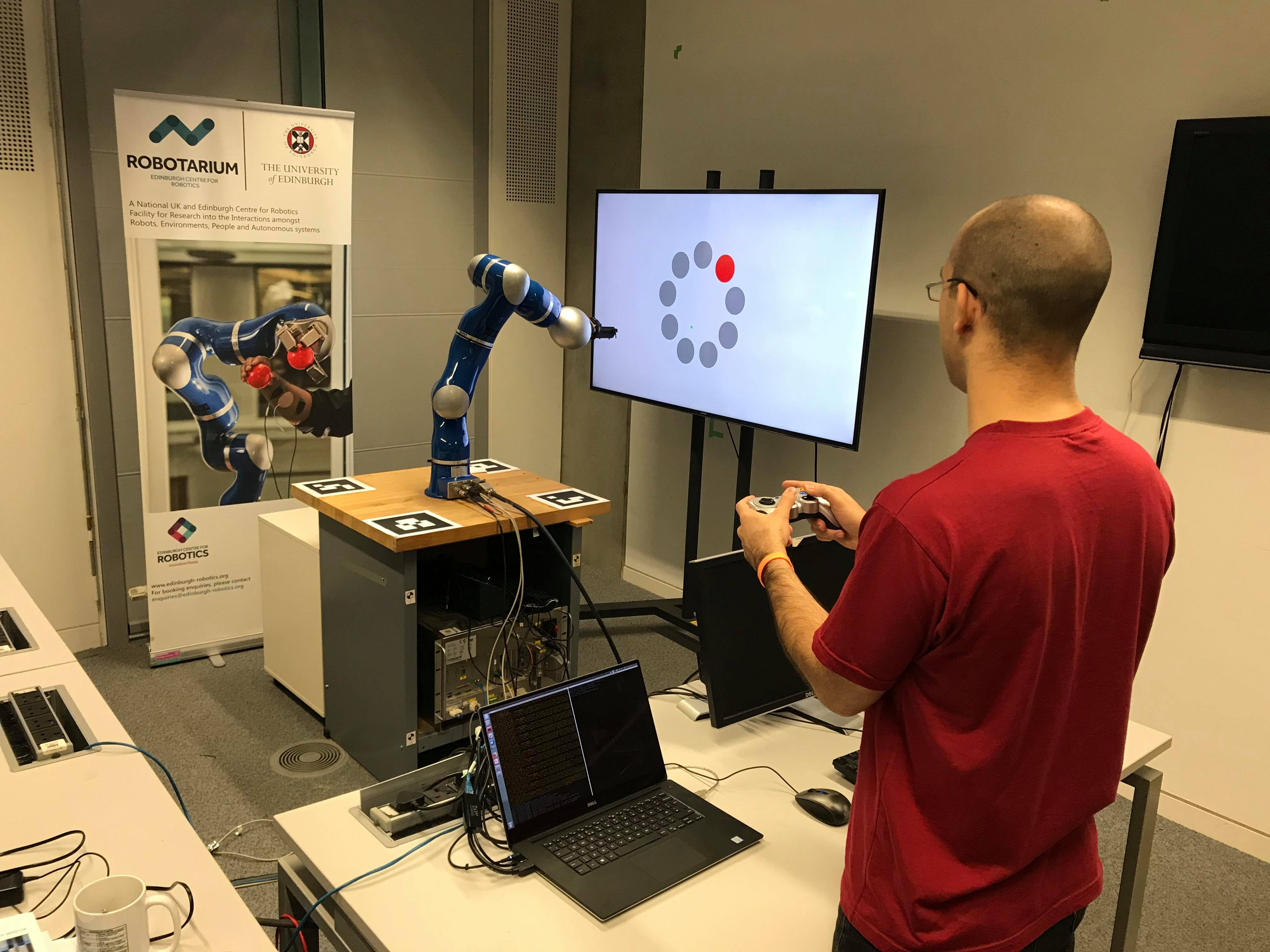}
    \label{fig:kuka-fitts}
  }
  ~
  \subfloat[Multi-directional task.]{
    \includegraphics[width=0.175\textwidth,trim={3cm 1cm 2.5cm 0.75cm},clip]{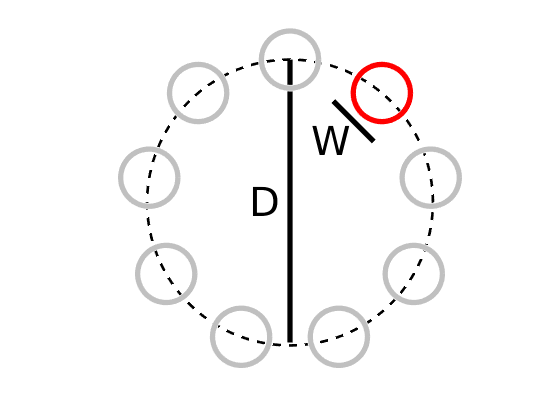}
    \label{fig:fitts-discs}
  }
  \\
  \subfloat[Target order of appearance for a single condition.]{
  \includegraphics[width=0.175\textwidth]{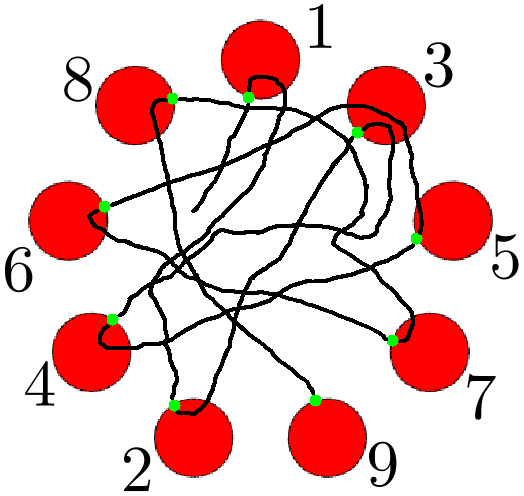}
  \label{fig:target-order}
  }
  ~
  \subfloat[Angle to the plane $\theta_t$ and standoff distance $\delta_t$ used to compute performance metrics.]{
    \includegraphics[width=0.275\textwidth]{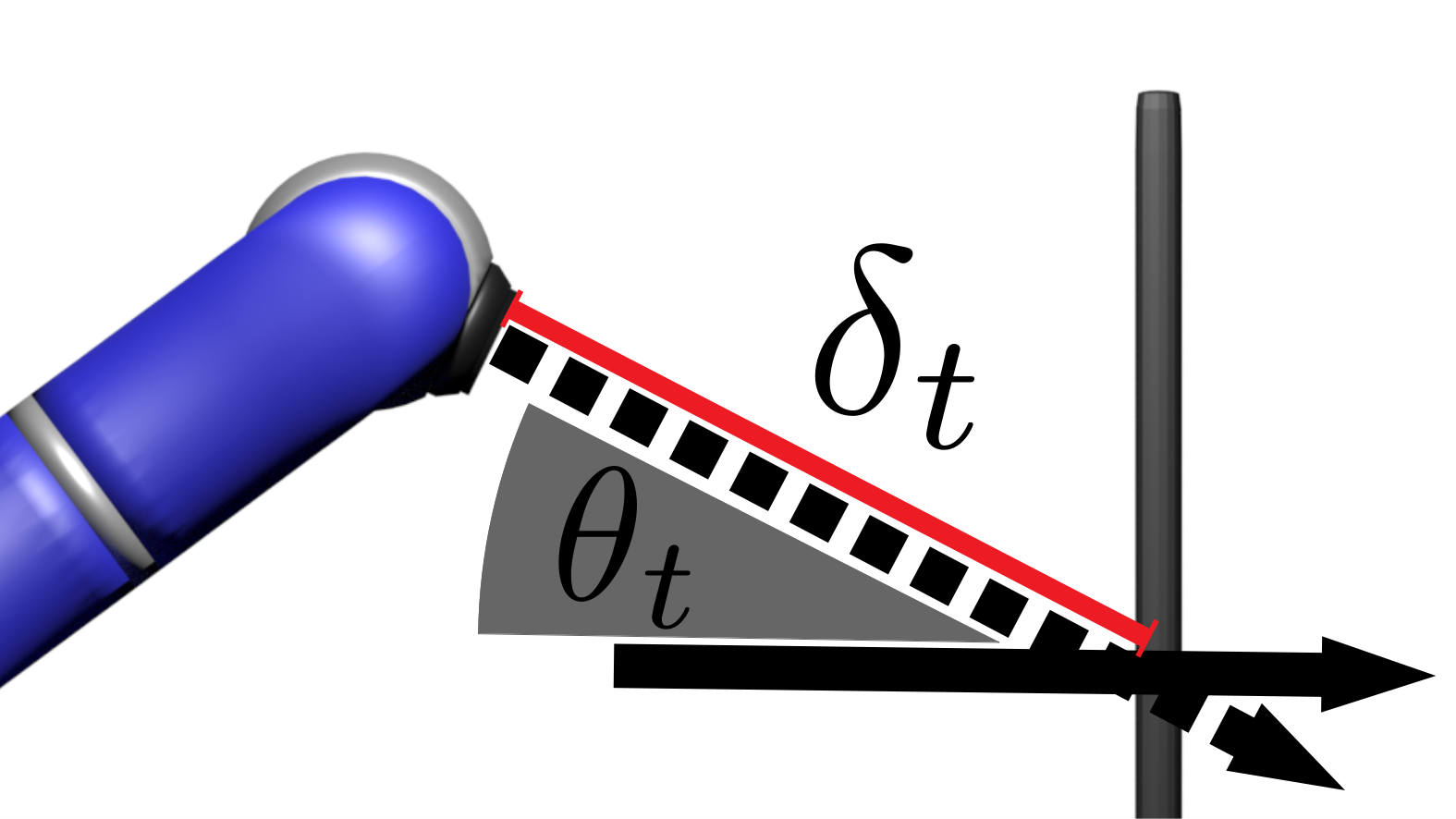}
    \label{fig:metrics}
  }
  \caption{Experimental setup. Whilst condition order of appearance is randomized, target order was always kept the same as in (c). The black line indicates an example path of the focus point during manipulation; the path was not shown to participants, only the focus point indicated by the green dot.}
  \label{fig:experimental-setup}
  \vspace{-1.5em}
\end{figure}

Fitts' established the information capacity of the human motor system by deriving a model for the the \textit{index of performance} $I_p$ (in \SI{}{\bit\per\second}) expressed by 
\begin{equation}
    I_p = I_d / T
\end{equation}
where $I_d$ is the \textit{index of difficulty} (in \SI{}{\bit}) and $T$ (in seconds) is the average movement time. The index of performance is a metric that quantifies task performance; higher values for $I_p$ indicate better performance. The index of difficulty is a metric that defines the task difficulty; higher values for $I_d$ imply the task is more difficult. Under the deterministic iterative-corrections model and by analogy with Shannons Theorem 17~\cite{Shannon63} a formula for $I_d$ is derived, see \cite{Crossman83} for details, given by
\begin{equation}
  \label{eq:index-of-difficulty}
  I_d = \log_2(2D/W)
\end{equation}
where $D$ is the distance to a target and $W$ is the width of the target, see Fig. \ref{fig:fitts-discs}. Define a \textit{condition} by the tuple $(W, D)$. Intuitively, a difficult task is when $I_d$ is large, thus $W$ is small and $D$ is large, and an easy task is when $I_d$ is small thus $W$  is large and $D$ is small.  

Operators in industry must maintain a number of motion constraints to achieve high task performance. Inspired from the concrete spraying task, we define two additional performance metrics that generalize Fitts' law. 

Define the \textit{angular length} $L_a$ as the total change in $\theta(t)$ for a complete move and expressed by 
\begin{equation}
  \label{eq:angular-length}
  L_a = \frac{1}{T}\int_0^T\theta_t dt
\end{equation}
where $\theta_t$ as in Fig. \ref{fig:metrics}. This quantity describes how well a user is able to maintain perpendicularity to the wall during transitioning between one target and the next. 

Define the \textit{delta length} denoted $L_\delta$ as the least absolute deviation in the standoff distance with respect to a given ideal over the duration of a complete move and expressed by 
\begin{equation}
    \label{eq:delta-length}
    L_\delta := \frac{1}{T}\int_0^T |\delta_t - \delta^*| dt
\end{equation}
where $\delta_t$ is the standoff distance as in Fig. \ref{fig:metrics} and $\delta^*$ is an ideal standoff distance. 

For each mode of teleoperation participants completed 17 conditions: 16 have been generated by scaling the condition values used in Fitts' original work~\cite{fitts1954information} while one has been hand-tuned to add spread to the $I_d$ values. The conditions used in this study are shown in \ref{tab:conditions} representing the range $I_d{\in}[1.3219, 5.9069]$.

\begin{table}
  \centering
  \caption{Condition values used in experiments.}\label{tab:conditions}
  \begin{tabularx}{0.4\textwidth}{ll}
    $W$ (m) & $D$ (m)\\\hline
    0.0125 & 0.0469, 0.0938, 0.1875, 0.3750\\
    0.0500 & 0.0938, 0.3750, 0.7500\\
    0.0250 & 0.0469, 0.0938, 0.1875, 0.7500\\
    0.1000 & 0.1875, 0.3750, 0.7500\\
    0.1500 & 0.1875, 0.3750, 0.7500\\
  \end{tabularx}
  \vspace{-1em}
\end{table}

\subsection{System description}

\begin{figure}[t]
    \centering
    \subfloat[Full joint space control.]{
        \includegraphics[width=0.23\textwidth, trim={3.5cm, 0.5cm, 3.5cm, 0cm}, clip]{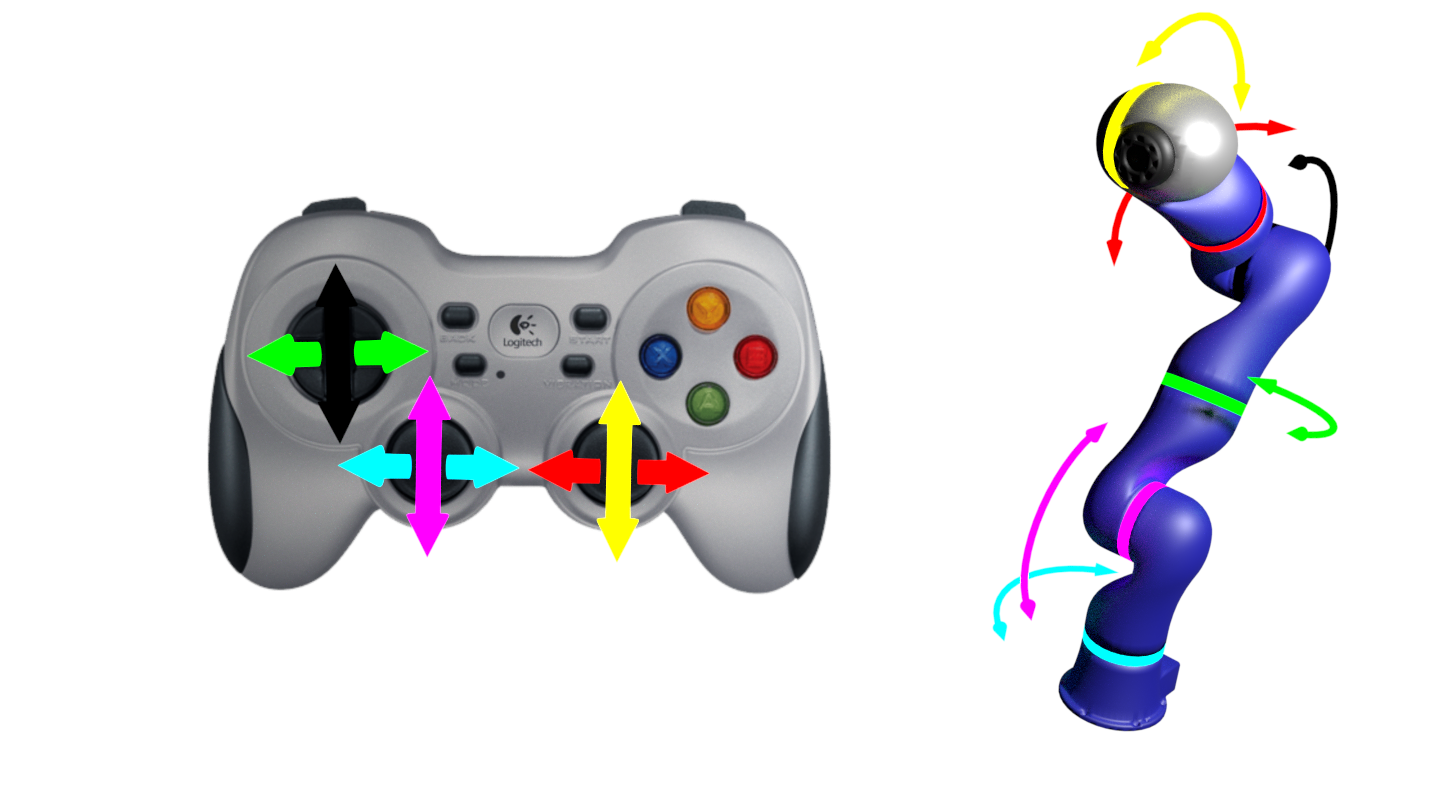}
    }
    ~
    \subfloat[Reduced joint space control.]{
        \includegraphics[width=0.23\textwidth, trim={3.5cm, 0.5cm, 3.5cm, 0cm}, clip]{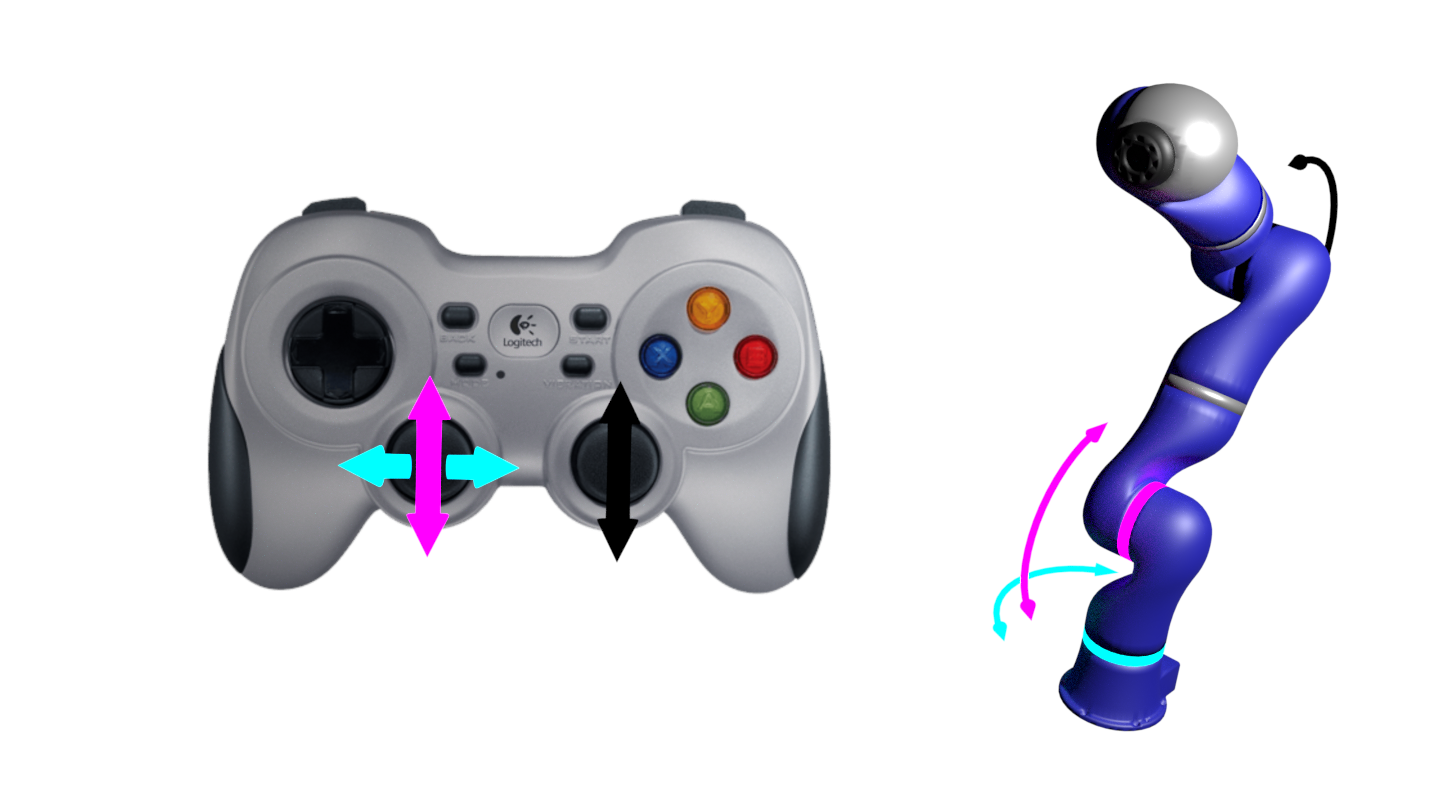}
    }
    \\
    \subfloat[Full task space control.]{
        \includegraphics[width=0.23\textwidth, trim={3.5cm, 0.5cm, 3.5cm, 0cm}, clip]{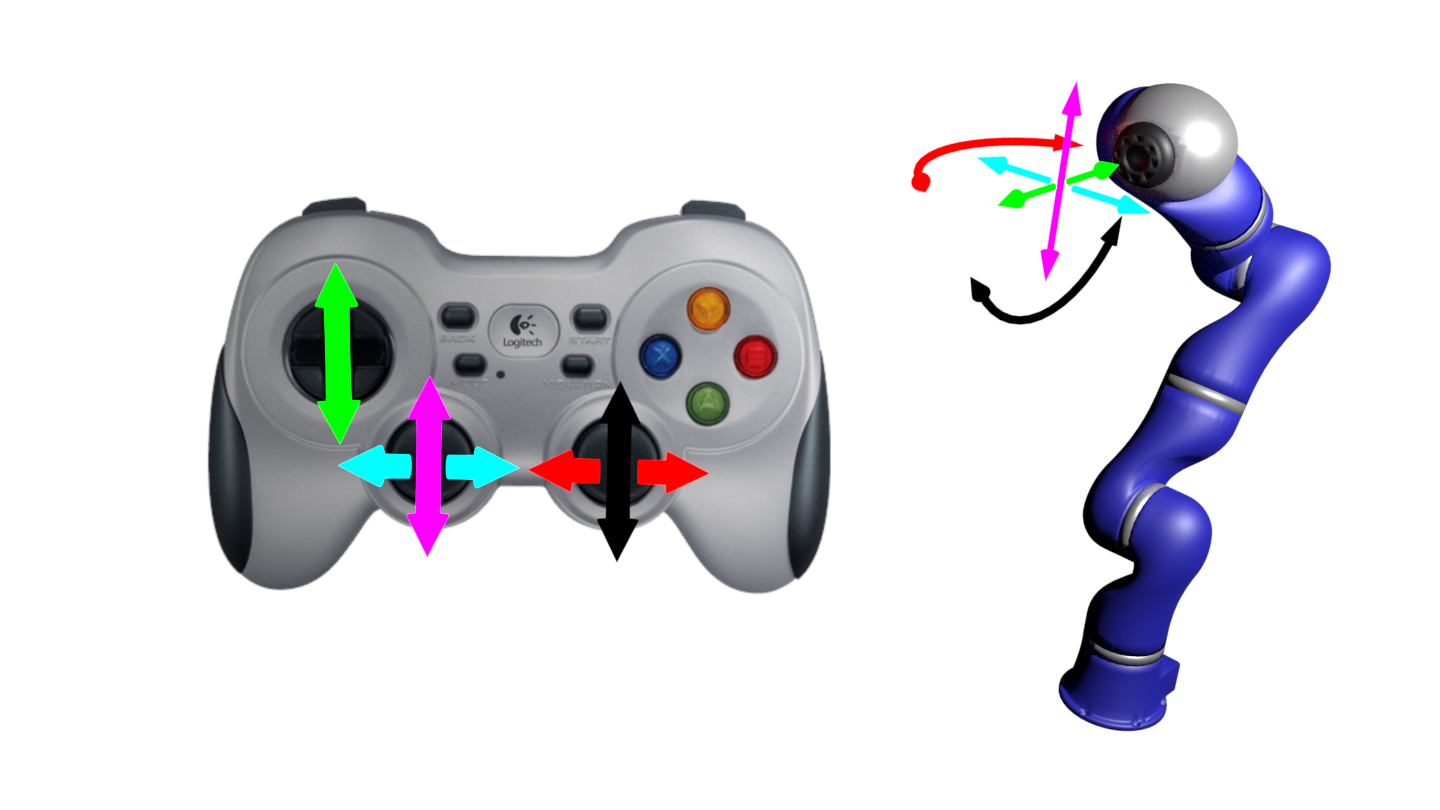}
    }
    ~
    \subfloat[Reduced task space control.]{
        \includegraphics[width=0.23\textwidth, trim={3.5cm, 0.5cm, 3.5cm, 0cm}, clip]{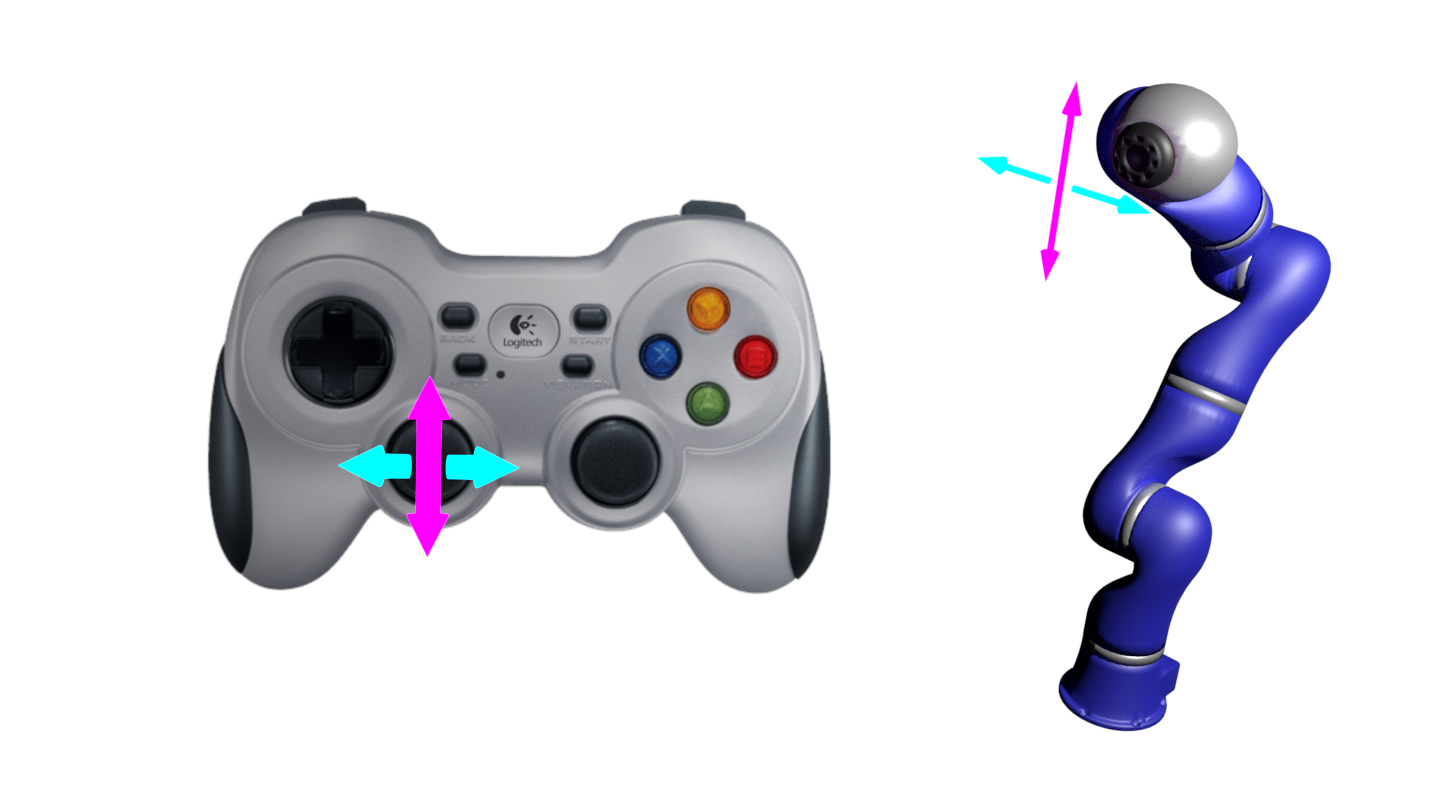}
    }
    \caption{Gamepad mappings for the modes of teleoperation implemented in experiments. Markings indicate where the human interacts with on the gamepad and where that interaction is perceived to be on the robot model.} 
    \label{fig:modes-of-teleoperation-kuka}
    \vspace{-2em}
\end{figure}

We have implemented each mode of teleoperation described in Sec. \ref{sec:modes-of-teleoperation} using a position control open-loop framework on a 7-DoF KUKA LWR robot arm. The operator interfaced with the system using an F710 Logitech gamepad, see Fig. \ref{fig:modes-of-teleoperation-kuka} for the mappings between the interface and robot for each mode. Each experiment was initiated by the user clicking a button on the gamepad. Inter-process communication is handled by the Robot Operating System (ROS)~\cite{Quigley2009} and modes requiring numerical optimization were computed using the Extensible Optimization Toolkit (EXOTica)~\cite{exotica}. Goal joint positions were streamed at \SI{100}{\hertz}. Targets and the focus point were displayed on a \SI{1.65}{\meter} cross-diagonal display placed specifically such that the transform between the robot base and the television screen was known. Targets were scaled to the real world and the position of the focus point was found using the robot forward kinematics.

An ideal standoff distance of $\delta^*=\SI{0.55}{\meter}$ was chosen, justified as follows: In industry, the MEYCO Oruga (Fig. \ref{fig:oruga}) has an approximate maximum reach of $r_O=\SI{6}{\meter}$ and the ideal standoff distance is $\delta^*_O=\SI{2}{\meter}$. The KUKA LWR arm has an approximate maximum reach of $r_K=\SI{1.5}{\meter}$. We thus find $\delta^*_K$ by balancing the ratios and adding a ten percent safety distance.
\vspace{-0.25em}
\subsection{Experimental protocol}
Participants were tasked with teleoperating a KUKA LWR robot arm to reach-and-point to a number of circular targets, indicated in red, arranged in a circle as in Fig. \ref{fig:fitts-discs}. Targets were displayed on a screen in a known position and orientation with respect to the robot base, see Fig. \ref{fig:kuka-fitts}. For each condition, targets were presented in the order as in Fig.~\ref{fig:target-order}. Conditions were randomized for every participant. The participants were instructed to use as many controllable dimensions available to them in order to manipulate the robot in such a way that completes the task as fast as possible while maintaining the angular and delta length constraints; i.e. simultaneously minimize $T$, $L_\alpha$, and $L_\delta$ to the best of their ability. 

Participants were allowed to move around the laboratory during the experiment, akin to concrete sprayer operators. For safety, they were not allowed within 1.5m of the robot.

Targets and conditions were presented to the participant in succession. As one target was deemed acquired the next immediately followed and as one condition was completed (i.e., one full cycle of targets in the order shown in Fig. \ref{fig:target-order}) the next immediately followed. Target acquisition is when the focus point (i.e., the point on the display screen the robot is pointing) comes into contact with the target. 

Experiments were completed for each mode of teleoperation. The order in which modes were presented to each participant was randomized to minimize skill-transfer. At the start of every mode the robot was reset to the same starting configuration. Following each mode participants were asked to fill out a questionnaire (Sec. \ref{sec:questionnaire}).

\subsection{Questionnaire}\label{sec:questionnaire}

Participants were asked to provide a rating on their speed perception, accuracy perception, fatigue, and the mental capacity for each mode. The questionnaire shown in Table \ref{tab:ques} was devised from the ISO 9241 standard and work by Douglas et al.~\cite{Douglas99}. Questions 1-5 use a 7-point scale and 6-7 are open-ended.

\begin{table}[H]
\caption{Questionnaire used in investigation.}\label{tab:ques}
    \begin{tabularx}{0.5\textwidth}{p{0.01em}p{20em}p{2.5em}p{2.5em}}
           &    Questions: \hfill Rating:                  & 1 & 7\\
        1) & The mental effort required for operation was \dotfill  & high & low\\
        2) & Accurate pointing was \dotfill & difficult  & easy\\
        3) & Operation speed was \dotfill  & fast & slow\\
        4) & Finger fatigue \dotfill  & high & none\\
        5) & Overall, the mode of teleoperation was \dotfill & difficult & easy \\
        6) & Did you have any trouble with this mode of teleoperation?\\
        7) & Do you have any comments in general about using this mode of teleoperation?
    \end{tabularx}
    \vspace{-1em}
\end{table}

\section{RESULTS}\label{sec:results}

\begin{figure*}
    \centering
    \includegraphics[width=0.8\textwidth, trim={3.5cm, 0.2cm, 4.35cm, 0.45cm}, clip]{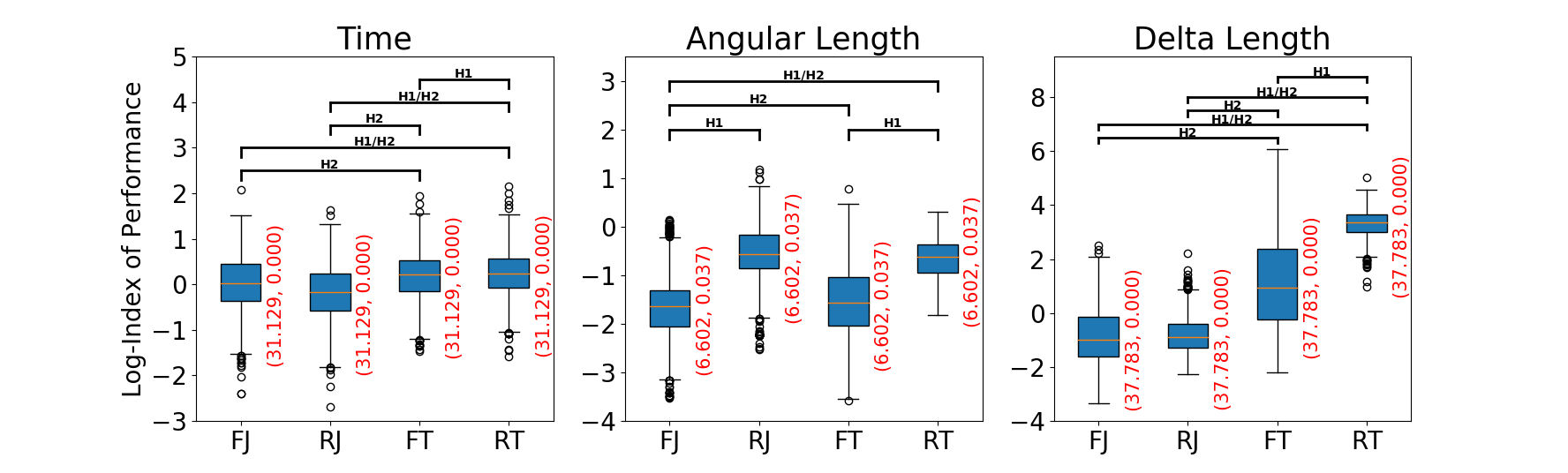}
    \caption{Log-Index of performance distribution results. Values in red indicate D'Agostino's $K^2$ normality test $(K^2, p)$. By means of a one-sided paired-sample t-test, pairs accepted under \hypref{hyp:first} and \hypref{hyp:second} ($\alpha=0.01$) are indicated above.}
    \label{fig:log-index-perf}
    \vspace{-0.5em}
\end{figure*}

\begin{figure*}[t]
    \centering
    \includegraphics[width=0.8\textwidth, trim={2.575cm, 0.4cm, 4cm, 0.675cm}, clip]{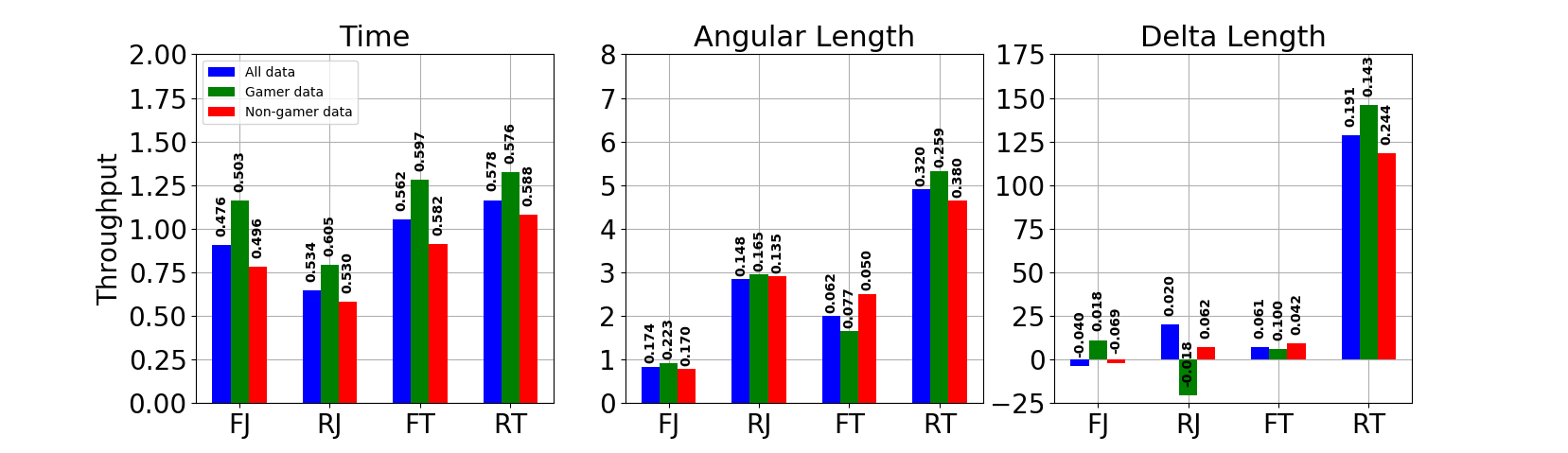}
    \caption{Throughput results that estimate the index of performance. We report the results for the full data set and the two subsets: gamer and non-gamer. We also provide the Pearson correlation coefficient above each bar.}
    \label{fig:throughput}
    \vspace{-1.5em}
\end{figure*}

\subsection{Low dimensional task space elicits high performance}\label{sec:results-low-dim-task-high-perf}
Data for 21 participants was collected for 17 conditions with 9 targets for each condition. In order to compute our performance metrics, we collected target acquisition timestamps and joint states. 

We have filtered the data based on the following. We ignore data for the first target of every condition since data collected when transitioning between conditions is not representative of skill in either condition. We also remove outliers that are outside 1.5 times the standard deviation of the task performance metrics. We have collected an extensive database of results constituting, in total, 8882 data points. Using each metric (time, angular length, and delta length) and the index of difficulty we compute an index of performance for every metric and for each point in our data set. The distributions of $\log(I_p)$ is shown in Fig. \ref{fig:log-index-perf}. Using this data we have been able to perform a one-sided paired-sampled t-test to test our hypotheses. 

Precisely the same conclusions for time and delta length metrics have been observed. We see \hypref{hyp:first} is accepted as expected for reducing control space dimensionality, except the FJ-RJ comparison. For the angular length \hypref{hyp:first} is accepted; this is unsurprising since RJ supports end-effector alignment. For cross-modes, e.g. FJ-RT, we see that \hypref{hyp:first} is accepted. 

When comparing between task space modes and joint space modes we see that \hypref{hyp:second} has been accepted. An interesting point to note that in the time metric and delta length metric \hypref{hyp:second} is accepted when comparing the RJ and FT modes. This suggests that low dimensional joint spaces in some cases elicit worse performance than higher dimensional task spaces. We postulate that the RJ mode may fight intuition resulting in the machine performing motions that do not match the operators innate model of the system dynamics.

\subsection{Habitual traits effect performance}
During preliminary investigations, it had been noted that participants who were known to play video games regularly seemed to have higher performance than those that did not. In our final experimental design we decided to include this comparison as part of our multivariate analysis. As described in Sec. \ref{sec:participant-selection} we asked each participant to indicate their gaming regularity; a participant is considered a \textit{gamer} if they gave a rating greater than or equal to 4, and a \textit{non-gamer} otherwise. Using this division we split our data set into two subsets. 

By the assumption that difficulty and performance are linearly correlated we are able to compare across the various modes of teleoperation using linear regression. In order to compare the general performance for each trait we use an estimate of \textit{throughput} \cite{fitts1954information}. The throughput is another indicator of overall performance. For the time metric, the throughput $Th$ is estimated by the reciprocal of a scalar $b$ where $a,b$ are regression parameters such that $a + bI_d = T$; so $Th = 1/b$. For the angular length and delta length, throughput for each has been estimated in the same fashion. Throughput estimates for our full data set and the gamer/non-gamer subsets are shown in Fig. \ref{fig:throughput}. 

Regarding time performance, in general, excluding the result for the RJ mode, a similar pattern is seen when compared to the results in Fig. \ref{fig:log-index-perf}; the RT mode has highest throughput in general and task space modes have higher throughput than joint space. For each mode of teleoperation we observe a higher throughput for gamers than the non-gamers. We noted that gamers generally seemed more familiar using two analog joysticks at once as opposed to the non-gamers. We posit, due to this ability, gamers were able to achieve faster times.

Regarding angular length performance, we see the general trend in performance as discussed in the previous section. Gamers have higher throughput than non-gamers for each mode except the FT mode. It should be noted that the correlation coefficients for these are generally low which renders these results as potentially spurious. Drawing conclusions from these results may be unreliable. The throughput results for the delta length unfortunately suffer from similar issues and so reliable conclusions cannot be made for these either. 

\subsection{Participants approve low dimensional task spaces}\label{sec:subjective-results}

\begin{table}[t]
  \centering
\caption{Results of the questionnaire and paired-sampled $t$-test ($\alpha = 0.05$). Bold indicates the one-sided hypothesis \hypref{hyp:third}/\hypref{hyp:fourth} is accepted. Note, since Q3 does not evaluate the users preference (highlighted in bold) indicates the result of a two-sided significance test.}\label{tab:ques}
  \begin{tabularx}{0.475\textwidth}{p{3em}p{5em}p{5em}p{5em}p{5em}}
    \hline
    Question 
    & FJ & RJ & FT & RT \\
    \hline
      1 & \FPeval{\result}{round(7-5.19,2)}\result~$\pm$ 1.33 & \FPeval{\result}{round(7-3.71,2)}\result~$\pm$ 1.59 & \FPeval{\result}{round(7-3.43,2)}\result~$\pm$ 1.57 & \FPeval{\result}{round(7-1.95,2)}\result~$\pm$ 1.36 \\ 
      2 & \FPeval{\result}{round(7-4.43,2)}\result~$\pm$ 1.66 & \FPeval{\result}{round(7-4.10,2)}\result~$\pm$ 1.73 & \FPeval{\result}{round(7-3.86,2)}\result~$\pm$ 1.39 & \FPeval{\result}{round(2.95,2)}\result~$\pm$ 1.69 \\ 
      3* & 4.62 $\pm$ 0.92 & 4.71 $\pm$ 1.06 & 3.81 $\pm$ 1.08 & 4.29 $\pm$ 1.15 \\
      4 & \FPeval{\result}{round(7-3.71,2)}\result~$\pm$ 2.00 & \FPeval{\result}{round(7-3.38,2)}\result~$\pm$ 1.88 & \FPeval{\result}{round(7-2.33,2)}\result~$\pm$ 1.43 & \FPeval{\result}{round(7-2.10,2)}\result~$\pm$ 1.37 \\ 
      5 & 3.10 $\pm$ 1.55 & 4.24 $\pm$ 1.26 & 4.86 $\pm$ 1.28 & 5.95 $\pm$ 1.07 \\
  \end{tabularx}
  \begin{tabularx}{0.475\textwidth}{p{3em}p{5em}p{5em}p{5em}p{5em}}
    \hline
      Question                 &     & \multicolumn{1}{l}{RJ} & \multicolumn{1}{l}{FT} & \multicolumn{1}{l}{RT}\\\hline
    \multirow{3}{*}{1} 
                       & FJ & \textbf{0.0048 (3)}            & \textbf{0.0009 (3)}            & \textbf{5.2e-7 (3/4)}             \\
                       & RJ &                   -     & 0.5867            & \textbf{0.0003 (3/4)}            \\
                       & FT &                  -      &  -                      & \textbf{0.0011 (3)}            \\ \hline
    \multirow{3}{*}{2} & FJ & 0.5538            & 0.1435            & \textbf{0.0281 (3/4)}            \\
                       & RJ &                -        & 0.6339            & \textbf{0.0104 (3/4)}            \\
                       & FT &                 -       &   -                     & 0.0727            \\ \hline
    \multirow{3}{*}{3*}
                       & FJ & 0.7245           & \textbf{0.0202}            & 0.2596            \\
                       & RJ &                -        & \textbf{0.0068}            & 0.1428            \\
                       & FT &                 -       &     -                   & 0.1158          \\ \hline
    \multirow{3}{*}{4} 
                       & FJ & 0.4907            & \textbf{0.0120 (3)}            & \textbf{0.0044 (3/4)}            \\
                       & RJ &           -             & \textbf{0.0100 (4)}            & \textbf{0.0005 (3/4)}            \\
                       & FT &            -            &     -                   & 0.3086            \\ \hline
    \multirow{3}{*}{5} 
                       & FJ & \textbf{0.0244 (3)}            & \textbf{0.0009 (3)}            & \textbf{3.3e-6 (3/4)}            \\
                       & RJ &             -           & 0.1198            & \textbf{1.0e-5 (3/4)}           \\
                       & FT &              -          &       -                 & \textbf{0.0047 (3)}            \\ \hline
  \end{tabularx}
  \vspace{-2em}
\end{table}

The results of the questionnaire are shown in Tab. \ref{tab:ques}. The mean and standard deviation of the responses on questions 1-6 are shown above a paired-sampled $t$-test to determine the responses' statistical significance. 

The results for question 1 show that participants felt the FJ mode required the highest amount of mental effort. There seems to be no statistical difference between the RJ and FT modes. Participants indicated the RT required the least amount of mental effort.

Overall, participants indicated the task space modes were the easiest to point accurately. There is no statistical difference between FJ and RJ modes both having ratings indicating that participants felt these were the most difficult to accurately point. The results of the $t$-test indicate we reject \hypref{hyp:third}/\hypref{hyp:fourth} when comparing FJ, RJ, and FT modes. However, whilst \hypref{hyp:third}/\hypref{hyp:fourth} are accepted when comparing between RT and the joint space modes,  \hypref{hyp:third} is rejected when comparing the task space modes. The p-value for the RJ-FT comparison is reasonably high and the FJ-FT comparison is higher than the p-value for the FT-RT comparison. Comparing the results of this question to the responses given for questions 6 and 7 (shown below), we suggest that, despite the computed p-value, accepting \hypref{hyp:third} for the FT-RT comparison has potential grounding as a conclusion. 

During our experiments, maximum joint velocities were reduced to a conservative range for safety. Question 3 attempted to ascertain whether the participants felt the robot motion was too slow or indeed too fast. Mean values indicate participants felt the operation speed was neither too fast nor too slow. There are not significant differences between the results apart from the FT mode compared to the joint space modes. 

The results for question 4 indicate the participants experienced the least finger fatigue for task space control modes with no statistical difference between the two. The joint space modes caused the most fatigue.

For question 5, participants rated the RT mode as the easiest to use and the FJ the most difficult. There is no significant difference between the FT and RJ modes.

We summarize the responses to questions 6 and 7. Selected are common statements made by participants.

\vspace{0.5em}
\noindent{\itshape\footnotesize Full joint mode:}\vspace{-0.65em}
\begin{table}[H]
    \centering
    \begin{tabularx}{0.475\textwidth}{p{0.5em}p{27.5em}}
    6) & ``\textit{Wrist joints felt slower than base joint.}''\\
       & ``\textit{Joint mappings felt inverted and was easy to get into strange configurations.}''\\
    7) & ``\textit{Hard to maintain constraints.}''\\
       & ``\textit{Only used two joints at a time.}''
    \end{tabularx}
\end{table}
\vspace{-1.5em}
\noindent{\itshape\footnotesize Reduced joint mode:}\vspace{-0.65em}
\begin{table}[H]
    \centering
    \begin{tabularx}{0.475\textwidth}{p{0.5em}p{27.5em}}
    6) & ``\textit{Attention was directed to focus point, not robot, making operation very difficult.}''\\
       & ``\textit{Often requiring to re-adjust.}''\\
    7) & ``\textit{Slightly better than full joint mode.}''\\
       & ``\textit{Easy for small $D$ values.}''
    \end{tabularx}
\end{table}
\vspace{-1.5em}
\noindent{\itshape\footnotesize Full task mode:}\vspace{-0.65em}
\begin{table}[H]
    \centering
    \begin{tabularx}{0.475\textwidth}{p{0.5em}p{27.5em}}
    6) & ``\textit{Unexpected motions at times.}''\\
       & ``\textit{Maintaining delta length was effectively impossible. Perhaps with more practice this mode would be more efficient.}''\\
    7) & ``\textit{Felt it was possible to go faster.}''\\
       & ``\textit{Favored over both joint space control modes; intuitive, mostly easy to use.}''
    \end{tabularx}
\end{table}
\vspace{-1.5em}
\noindent{\itshape\footnotesize Reduced task control mode:}\vspace{-0.65em}
\begin{table}[H]
    \centering
    \begin{tabularx}{0.475\textwidth}{p{0.5em}p{27.5em}}
    6) & ``\textit{Some unexpected motions.}''\\
       & ``\textit{Sometimes felt too slow.}''\\
    7) & ``\textit{Very intuitive and easy to use.}''
    \end{tabularx}
    \vspace{-1em}
\end{table}

The responses generally correlate with the answers to questions (1)-(5). The mappings for both joint modes were identified as inverted by many of the participants. The direction the joint moved under these modes was matched with the joystick direction. For example, for the FJ mode, pushing forward on the left joystick moved joint 2 in the direction such that the end-effector and focus point moved downwards. Participants felt the robot end-effector and focus point should instead move upwards. An issue for some participants for the task space modes is that the robot would make unexpected motions in certain configurations. The source of this issue is that at times a target configuration was computed requiring a joint velocity surpassing the maximum; an unfortunate consequence of unconstrained optimization. Whilst these issues were observed at times they did not render the task impossible to complete.


\section{Conclusions}
In this paper, we developed, implemented, and carried out a study to determine which modes of teleoperation elicit high task performance for unskilled human operators on a task inspired by concrete spraying in industry. A Fitts' law paradigm was used to quantify difficulty and educe the performance of each mode. We have generalized Fitts' law for two additional performance metrics. An extensive data set was collected from an experiment consisting of 21 participants. The results and analysis performed support several conclusions regarding control, sub-task allocation between human and autonomy, and how habitual traits can effect performance. 

A reduced task space control mode with has been shown to outperform all three other control modes with regards to the index of performance over three performance metrics; time, angular length, and delta length. The results of the questionnaire support this conclusion as the RT mode was generally favored the highest. The RT mode is the closest model that directly regulates the performance parameters. Given that our hypotheses were were accepted this opens new avenues for shared control design. Based on these results, we posit the contention that there is a positive correlation between cognitive load and number of human controllable dimensions for task space control modes. Both joint space modes were not favored by participants in this study with some noting that the reduced joint space control felt like having the ``\textit{worst of both worlds}'' with regards to joint space and task space control.

This work assumed unskilled participants and so the long-term learning effects on performance are not considered here. In future work, we intend to study the effect of learning curve in order to predict average learning time. The ability of the participants classed as gamers to achieve higher performance than non-gamers may be a consequence of their familiarity with the game pad controller used as the interface in our experiments. Future investigations will take this into account the effect the interface used by comparing other interfaces such as a joystick, 6-DoF space mouse, a combination of computer mouse and keyboard. Metrics quantifying cognitive load will be explored and compared against the number of controllable dimensions in-order to investigate our proposition of a positive correlation between the two quantities. 

This study indicates the mode specification highly impacts the design and performance of shared control systems. We intend to use the knowledge acquired here to inform the development of new formulations for shared autonomous and collaborative methods that adapt dynamic motion constraints on-the-fly using multi-modal sensory data.


\bibliography{bib}
\bibliographystyle{IEEEtran}

\end{document}